\documentclass[10pt,letterpaper]{article}
\usepackage[top=0.85in,left=1.0in,footskip=0.75in]{geometry}

\usepackage{changepage}

\usepackage[utf8]{inputenc}

\usepackage{textcomp,marvosym}

\usepackage{fixltx2e}

\usepackage{amsmath,amssymb}

\usepackage{cite}

\usepackage{nameref,hyperref}

\usepackage{microtype}
\DisableLigatures[f]{encoding = *, family = * }

\usepackage{rotating}

\usepackage{ragged2e} 

\raggedright
\usepackage[aboveskip=1pt,labelfont=bf,labelsep=period,justification=raggedright,singlelinecheck=off]{caption}

\makeatletter
\renewcommand{\@biblabel}[1]{\quad#1.}
\makeatother

\date{}

\graphicspath{{figures/}} 
\begin{document}

\begin{flushleft}
{\Large
\textbf\newline{Learning by Stimulation Avoidance: \\A Principle to Control Spiking Neural Networks Dynamics}
}
\newline
\\
Lana Sinapayen\textsuperscript{1 \Yinyang* \textcurrency },
Atsushi Masumori\textsuperscript{1\P},
Takashi Ikegami\textsuperscript{1\ddag}
\\
\bigskip
\bf{1} The University of Tokyo, Ikegami Laboratory, Tokyo, Japan
\\
\bigskip

%
%
\Yinyang Designed research, produced and analysed data and wrote paper.

\ddag  Designed research, contributed analytic tools and revised paper.

\P Contributed simulation ideas.

\textcurrency Current address: The University of Tokyo, Graduate School of Arts and Sciences, Department of General Systems Studies, Tokyo, Japan

* lana@sacral.c.u-tokyo.ac.jp

\end{flushleft}

\justifying 

\section*{Abstract}
Learning based on networks of real neurons, and by extension biologically inspired models of neural networks, has yet to find general learning rules leading to widespread applications.
In this paper, we argue for the existence of a principle allowing to steer the dynamics of a biologically inspired neural network. Using carefully timed external stimulation, the network can be driven towards a desired dynamical state. We term this principle ``Learning by Stimulation Avoidance'' (LSA). We demonstrate through simulation that the minimal sufficient conditions leading to LSA in artificial networks are also sufficient to reproduce learning results similar to those obtained in biological neurons by Shahaf and Marom~\cite{ShahafLearning}. We examine the mechanism's basic dynamics in a reduced network, and demonstrate how it scales up to a network of 100 neurons. 
We show that LSA has a higher explanatory power than existing hypotheses about the response of biological neural networks to external simulation, and can be used as a learning rule for an embodied application: learning of wall avoidance by a simulated robot. The surge in popularity of artificial neural networks is mostly directed to disembodied models of neurons with biologically irrelevant dynamics: to the authors' knowledge, this is the first work demonstrating sensory-motor learning with random spiking networks through pure Hebbian learning.

\section*{Author Summary}
Networks of spiking neurons are currently the model that most closely reproduces real neuron's dynamics~\cite{izhWhichModel}. Theoretical approaches have shown that spiking models have a tremendous potential for learning and processing information~\cite{maass1997networks}, but efforts are still ongoing to use this potential in mainstream applications. By finding general learning rules governing the dynamics of spiking networks, we get closer to this goal. In this paper, we propose such a biologically plausible rule (LSA), explain how it influences the dynamics of the network, demonstrate how to use it in practice to obtain desired dynamics from initially random networks and finally, show that it can be used to obtained desired behaviour in a simulated robot. Far from being constrained to simulated models, LSA also offers an interpretation to so far unexplained experimental results obtained with in vitro networks.


\section*{Introduction}

In two papers published in 2001 and 2002, Shahaf and Marom conduct experiments with a training method that drives rats' cortical neurons cultivated in vitro to learn given tasks~\cite{ShahafLearning, MaromDevelopment}. They show that stimulating the network with a focal current and removing that stimulation when a desired behaviour is executed is sufficient to strengthen said behaviour. By the end of training, the behaviour is obtained reliably and quickly in response to the stimulation.
More specifically, networks learn to increase the firing rate of a group of neurons (output neurons) inside a time window of 50~ms, in response to an external electric stimulation applied to an other part of the network (input neurons). This result is powerful, first due to its generality: the network is initially random, the input and output zones' size and position are chosen by the experimenter, as well as the output's time window and the desired output pattern. A second attractive feature of the experiment is the simplicity of the training method. To obtain learning in the network, Shahaf and Marom repeat the following two steps: (1)~Apply a focal electrical stimulation to the network. (2)~When the desired behavior appears, remove the stimulation.

At first the desired output seldom appears in the required time window, but after several training cycles (repeating steps (1) and (2)), the output is reliably obtained.
Marom explains these results by invoking the Stimulus Regulation Principle (SRP, from~\cite{HullSRP, GuthrieSRP}). At the low level of a neural network, the SRP postulates that stimulation drives the network to ``try out'' different topologies by modifying neuronal connections (``modifiability''), and that removing the stimulus simply freezes the network in its last configuration (``stability''). The SRP explicitly postulates that no strengthening of neural connections occurs as a result of stimulus removal.

The generality of the results obtained by Shahaf and Marom suggests that this form of learning must be a crucial and very basic property of biological neural networks. But the SRP does not entirely explain the experimental results. Why are several training cycles necessary if ``stability" guarantees that the configuration of the network is preserved after stopping the stimulation? How does ``modifiability" not conflict with the idea of learning, if we cannot prevent the ``good'' topology to be modified by the stimulation at each new training cycle?

We propose a different explanatory mechanism: the principle of Learning By Stimulation Avoidance (LSA,~\cite{LanaECAL2015, MasumoriECAL2015}). LSA is an emergent property of spiking networks coupled to Hebbian rule~\cite{hebbianLearning}  and external stimulation. LSA states that these networks will reliably learn to exhibit firing patterns that lead to the removal of an external stimulation (positive LSA), and avoid exhibiting firing patterns that lead to the application of an external stimulation (negative LSA).

In opposition to the SRP, LSA does not postulate that stimulus intensity is the major drive for changes in the network, but rather that the timing of the stimulation relative to network activity is crucial. LSA relies entirely on time dependent strengthening and weakening of neural connections. In addition, LSA proposes an explanatory mechanism for synaptic pruning, which is not covered by the SRP.

Shahaf postulates that the SRP might not be at work in ``real brains''. Indeed, while SRP has not yet been found to take place in the brain, the fundamental rule from which LSA emerges, Spike-Timing Dependant Plasticity (STDP), has been found in both in vivo and in vitro networks. STDP is a Hebbian learning rule so fundamental that it has been consistently found in the brains of a wide range of species, from insects to humans~\cite{CaporaleSTDP, PerceptionSTDP,SongSTDP}. STDP causes changes in the synaptic weight between two firing neurons depending on the timing of their activity: if the presynaptic neuron fires within 20~ms before the postsynaptic neuron, the synaptic weight increases; if the presynaptic neuron fires within 20~ms aftter the postsynaptic neuron , the synaptic weight decreases.

Although STDP occurs at neuronal level, it has very direct consequences on the sensory-motor coupling of animals with the environment. In vitro and in vivo experiments based on STDP can reliably enhance sensory coupling~\cite{STDPEnhance}, decrease it~\cite{JacobSTDP}, and these bidirectional changes can even be combined to create receptive fields in sensory neurons~\cite{VisualCortex, CatMap}.

\begin{figure}
\includegraphics[width=1.0\linewidth]{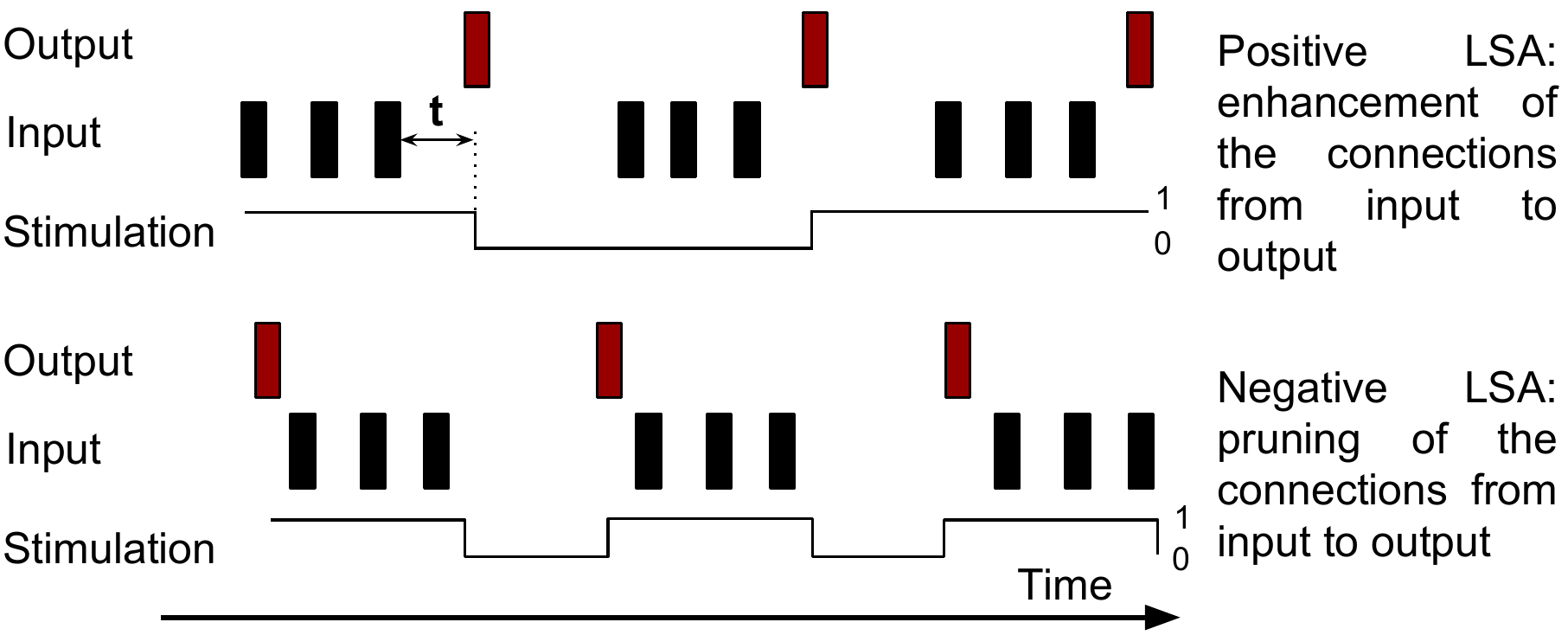}
\caption{{\bf Using external stimulation to obtain LSA between neurons when the synaptic changes are mediated by STDP.} Stopping the stimulation when the output neuron fires leads to the necessary STDP condition for increasing the weight (positive LSA): ``input last fired just before output''. It is interesting to note that without the firing latency \textbf{t}, this effect of STDP would not be possible. Starting the stimulation when the output fires leads to the necessary condition for decreasing the weight (negative LSA): ``output last fired just before input''.}
\label{fig:lsaDyn}
\end{figure}

In all these studies the reaction of cortical neurons to stimuli is gradually enhanced when they are made to fire within 20~ms after the end of the stimulation, and inhibited when they are made to fire within 20~ms after the start of the stimulation. In this paper, we show that these are the only conditions required for the emergence of both positive LSA and negative LSA between pre- and post-synaptic neurons, as illustrated in Fig.~\ref{fig:lsaDyn}. Therefore, although STDP is a rule that operates at the scale of one neuron, LSA can be expected to emerge at network level in real brains as well as it emerges in artificial networks. LSA at a network level requires an additional condition that is burst suppression. In this paper, we have tested two mechanisms. One is simply adding white noise to all neurons and the other one is to use a short term plasticity rule (STP).

The structure of the paper is as follows: the model is presented in Section~\ref{sec:model}; we show that the conditions necessary to obtain LSA are sufficient reproduce biological results in Section~\ref{sec:LSAbio} and study the dynamics of LSA in a minimal network of 3 neurons in Section~\ref{sec:LSAminimal}. We then show that LSA supports hypotheses to explain biological mechanisms that are not covered by the SRP (Section~\ref{sec:LSAvsSRP}) and implement a simple embodied application using LSA as the sole learning mechanism in Section~\ref{sec:LSArobot}. In Section~\ref{sec:Param} we explore the effect of parameter changes on the learning performance of the network.

\section{Model}
\label{sec:model}
\section*{Network model}

\begin{figure}
	\includegraphics[width=1.0\linewidth]{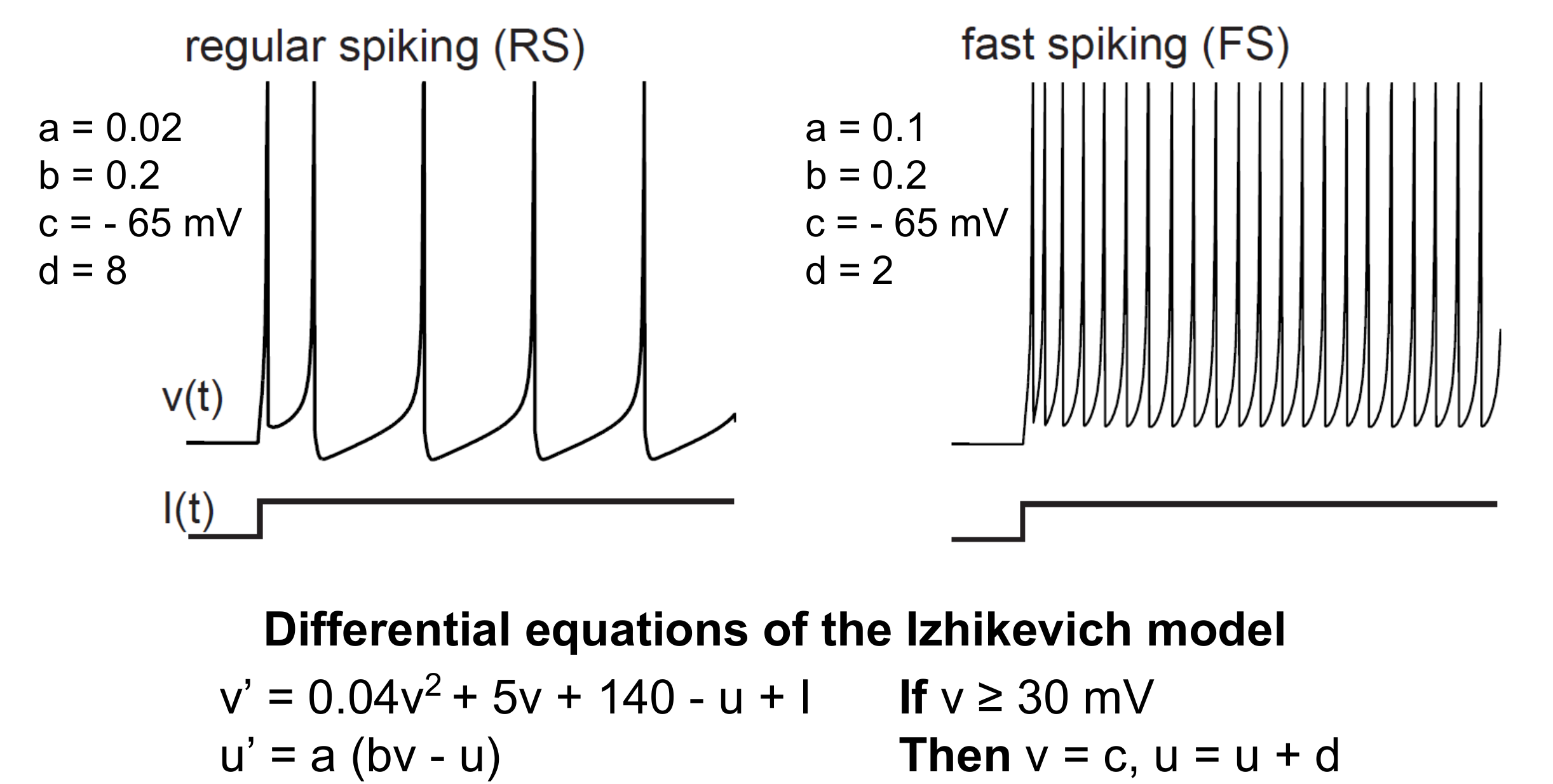}
\caption{{\bf Equations and dynamics of regular spiking and fast spiking neurons simulated with the Izhikevich model.}
	Equations and dynamics of regular spiking and fast spiking neurons simulated with the Izhikevich model. The spiking figures are reproduced with permission from www.izhikevich.com. (Electronic version of the figure and reproduction permissions are freely available at www.izhikevich.com)}
	\label{fig:neuronDynamics}
	\end{figure}

We use the model of spiking neuron devised by Izhikevich~\cite{izhSimpleModel} to simulate excitatory neurons (regular spiking neurons) and inhibitory neurons (fast spiking neurons) with a simulation time step of 1~ms. The equations of the neural model and the resulting dynamics are shown in Fig.~\ref{fig:neuronDynamics}.
In the experiments of Section~\ref{sec:LSAbio} and Section~\ref{sec:LSArobot}, we simulate fully connected networks of 100 neurons (self-connections are forbidden) with 80 excitatory and 20 inhibitory neurons. This ratio of 20\% of inhibitory neurons is standard in simulations~\cite{izhSimpleModel, izhWhichModel} and close to real biological values (15\%,~\cite{GABAratio}). The initial weights are random (uniform distribution: $0<w<5$ for excitatory neurons, $-5<w<0$ for inhibitory neurons). The maximum weight is fixed to 10. There are no delays in signal transmission between neurons. The neurons receive three kinds of input: (1) Gaussian noise $m$ with a standard deviation $\sigma = 3$~mV, representing noisy thalamic input. (2)~External stimulation $e$ with a value of 0 or 1 mV and a frequency of 1000~Hz. The timing of the stimulation depends on the experiment. (3) Stimulation from other neurons: when a neuron $a$ spikes, the weight $w_{a,b}$ is added as an input for neuron $b$.
All these inputs are added for each neuron $n_i$ at each iteration as:
	
\begin{eqnarray}
I_i = I_i^*+ e_i + m_i \;.
\end{eqnarray}

\begin{eqnarray}
\label{eq:neuronInput}
I_i^*= \sum_{j=0}^{n}w_{j,i} \times f_j\;,
f_j = 
\begin{cases}
    1 , & \text{if } n_j \text{ is firing}\\
    0, & \text{otherwise}.
\end{cases}
\end{eqnarray}

In the experiments with a reduced network (Section~\ref{sec:LSAminimal}), we simulate only 3 neurons, all excitatory. The initial weights are fixed to 5, except if noted otherwise. In the experiments of Section~\ref{sec:LSAvsSRP}, burst suppression is obtained in the 100-neuron network by reducing the number of connections to obtain a sparsely connected network: each neuron has 20 random connections to other neurons (uniform distribution, $0<w<10$), a high maximum weight  of 50, high external input $e=10$~mV and high noise $\sigma =5$~mV\footnote{Variations in the number of connections and the strength variance are examined in Section~\ref{sec:Param} }.

\section*{Plasticity model}

\begin{figure}
	\includegraphics[width=0.9\linewidth]{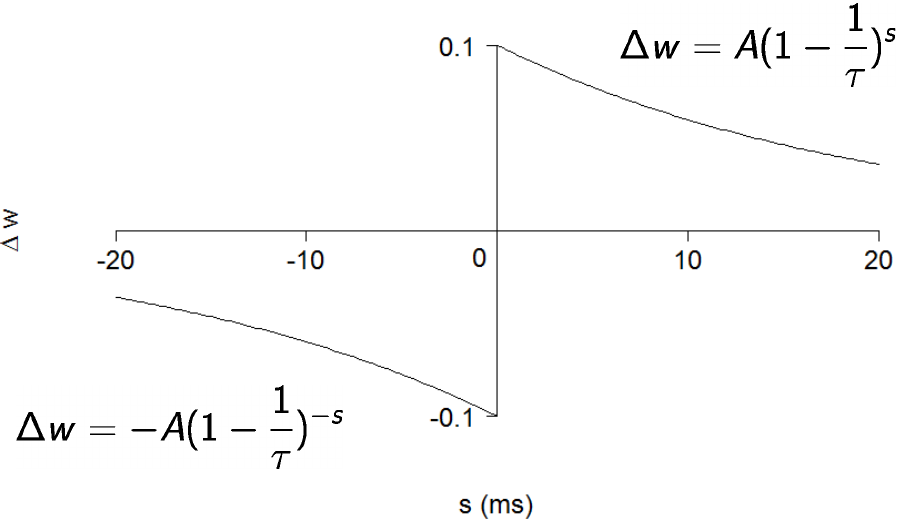}
	\caption{{\bf The Spike-Timing Dependent Plasticity (STDP) function} governing the weight variation $\Delta w$ of the synapse from neuron $a$ to neuron $b$ depending on the relative spike timing ${s= t_b-t_a}$. $A = 0.1$; $\tau=20$ ms.}
	\label{fig:stdp}
	\end{figure}

We add synaptic plasticity to all networks in the form of STDP as proposed in \cite{bushSTDP}.  STDP is applied only between excitatory neurons; other connections keep their initial weight during all the simulation. We use additive STDP: Fig.~\ref{fig:stdp} shows the variation of weight $\Delta w$ for a synapse going from neuron $a$ to neuron $b$. As shown on the figure, $\Delta w$ is negative if $b$ fires first, and positive $a$ fires first. 
The total weight $w$ varies as:

\begin{eqnarray}
w_t = w_{t-1}+ \Delta w \;.
\label{eq:wvariation}
\end{eqnarray}

We fix a maximum value to the weight: if $w > w_{max}$,  $w$ is reset to $w_{max}$. In the experiments with 100-neurons networks, we also apply a decay function to all the weights in the network. The decay function is applied at each iteration $t$ as:
	
\begin{eqnarray}
\forall w_t,\; w_{t+1} = (1 - \mu) w_t
\label{eq:decay}
\end{eqnarray}

\begin{eqnarray}
\mu  = 5 \times 10^{-7}\;.
\end{eqnarray}
	
In the experiments of Section~\ref{sec:LSArobot}, burst suppression is obtained by adding a phenomenological model of Short Term Plasticity~(STP, \cite{STP}) to the network, as a way to suppress bursts while keeping the network fully connected. STP is a reversible plasticity rule that decreases the intensity of neuronal spikes if they are too close in time, preventing the network to enter a state of global synchronized activity:

\begin{eqnarray}
w^*_{i,j} = u xw_{i,j}
\end{eqnarray}

\begin{eqnarray}
\frac{dx}{dt} = \frac{1 - x }{\tau_d} - uxf_i
\end{eqnarray}

\begin{eqnarray}
\frac{du}{dt} = \frac{U - u }{\tau_f} + U(1-u)f_i
\end{eqnarray}

STP acts as a short term reversible factor on the original synaptic weight, with the side effect of preventing global bursting of the network. Eq.~\ref{eq:neuronInput} becomes

\begin{eqnarray}
I_i^*= \sum_{j=0}^{n}w^*_{j,i} \times f_j\;.
\end{eqnarray}

\section*{Robot}

In Section~\ref{sec:LSArobot}, we simulate a simple robot moving inside an arena surrounded by walls. The arena is a square of size 1000~px (pixels). The robot is a 25~px radius circle, constantly moving at 1~px/ms except in case of collision with a wall. The robot has two distance sensors oriented respectively at $\pi/4$ and $-\pi/4$ from the front direction of the robot. The sensors have a range of 80~px; they are activated when the robot is at less than 80~px from a wall, on the direction supported by each sensor's orientation. Two input zones in the network (10 neurons each) receive input in mV from the sensors as follows:

\begin{eqnarray}
output = sensitivity / distance
\end{eqnarray}

The $sensitivity$ of the sensors is fixed at a constant value for the duration of each experiment.
For simplicity, the robot's steering is non-differential. It is controlled by the spikes of two output zones in the network (10 neurons each). For each spike in the left output zone, the robots steers
$\pi/6$ radian (to the left); for each spike in the right output zone, the robots steers $-\pi/6$ radian (to the right).

\section{Results}
\subsection{LSA is Sufficient to Explain Biological Results}
\label{sec:LSAbio}

In \cite{LanaECAL2015} we showed that a simulated random spiking network built from \cite{izhWhichModel, bretteSpikingModels} combined to STDP could be driven to learn desired output patterns using a training method similar to that of Shahaf et al. Shahaf shows that his training protocol can reduce the response time of a network. The response time is defined as the delay between the the application of the stimulation and the observation of a desired output from the network. In his first series of experiments (``simple learning'' experiments), the desired output is defined by the fulfilment of one condition, Condition~1: the electrical activity must increase in a chosen Output~Zone~A. This is the experiment we reproduced in \cite{LanaECAL2015}, demonstrating that this learning behaviour is a direct effect of STDP and is captured by the principle of LSA: firing patterns leading to the removal of external stimulation are strengthened, firing patterns that lead to the application of an external stimulation are avoided. 

In this section we show that the same methods are sufficient to obtain results similar to the second series of experiments performed by Shahaf (``selective learning" experiments), in which the desired output is the simultaneous fulfilment of Condition 1 as defined before and Condition~2:  a different output zone (Output~Zone~ B) must not exhibit enhanced electrical activity. When both conditions are fulfilled, the result is called selective learning because only Output Zone A must learn to increase its activity inside the time window, while Output Zone B must not change its activity.

We reproduce the experiment as follows. In a network of 100 neurons, a group of 10 excitatory neurons are stimulated with 1~mV at a frequency of 1,000~Hz. Two different groups of 10 neurons are monitored (Output Zone A and Output Zone B). We define the desired output pattern as: 4 neurons or more firing in Output~Zone~A (Condition~1), and less than 4 neurons firing in Output~Zone~B (Condition~2). Both conditions must be fulfilled simultaneously, where simultaneously means at the same millisecond. We stop the external stimulation as soon as the desired output is observed. If the desired output is not observed after 10,000~ms of stimulation, the stimulation is also stopped. After a random delay of 1,000 to 2,000~ms, the stimulation starts again.
In addition, each neuron is stimulated with noisy input and the weights are subject to slow decay as indicated in the Model section.

This paper provides a theoretical explanation to Shahaf's in vitro experiment by simulating spiking neurons. There are important differences: the stimulation frequency (Shahaf uses lower frequencies), its intensity (this parameter is unknown in Shahaf's experiment) and the time window for the output (in Shahaf's results the activity of Output Zone A is arguably higher even outside of the selected output window). We also use a fully connected network, while the biological network grown in vitro is likely to be sparsely connected \cite{inVitroTopology}.

\begin{figure}
\includegraphics[width=1.0\linewidth]{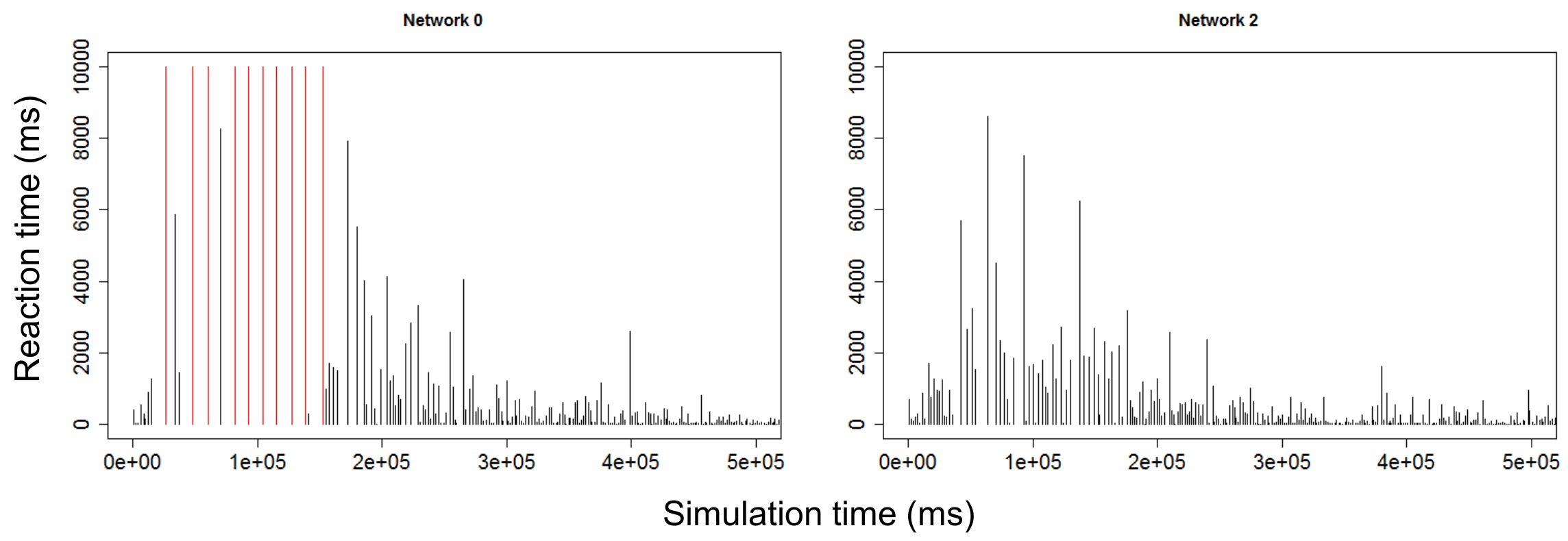}
\caption{{\bf Evolution of the reaction times of 2 successful neural networks at the selective learning task.} 20 networks were simulated, 18 of which successfully learned the task (Table~\ref{tab:stats1}). Red lines represent training cycles that ended without the network producing the expected correct output. A learning curve is clearly visible.}
\label{fig:repro}
\end{figure}

Despite these differences, we obtain results comparable to those of Shahaf: the reaction time, initially random, becomes shorter with training (Fig.~\ref{fig:repro}). We also perform the experiment with no stimulation at all and find a success rate of 0\%; the statistics of the selective learning experiment are summarized in~Table~\ref{tab:stats1}.

\begin{table}[t]
\caption{
{\bf Statistical performance of the network}}
\begin{tabular}{ l  c  c  c}
Condition & Success rate & Learning time & Final reaction time\\
\hline\\\
Selective learning & 90\% & 187 $\pm$ 16 s & 389 $\pm$ 54  ms\\ 
\hline\\\
No external stim. & 0\% & -- & -- \\ 
\hline
\end{tabular}
\begin{flushleft} The task is learned (learning time) when the reaction time of the network reaches a value inferior to 4,000 ms and keeps under this limit. The success rate is the percentage of networks that successfully learned the task in 400,000~ms or less (N = 20 networks per condition). The final reaction time is calculated for successful networks after learning. Standard error is indicated.
\end{flushleft}
\label{tab:stats1}
\end{table}

As shown by these results, the network exhibits selective learning as defined by Shahaf. But we also find that despite a success rate of 90\% at exhibiting the desired firing pattern, both the firing rates of Output Zone A and Output Zone B increase in equivalent proportions: the two output zones fire at the same rate but in a desynchronized way (see also Fig.~\ref{fig:steering}-b). Although data about firing rates is not specifically discussed in the paper, Shahaf himself reports in his experiment that only half of the in-vitro networks succeeded at selective learning, while all succeeded at the 
``simple learning" task. Our hypothesis is that bursts are detrimental to learning \cite{Bursts} and explain the difficulty of obtaining selective learning. If this hypothesis is true, burst suppression should improve learning.
Before discussing burst suppression, in the next section we quickly survey the dynamics of LSA in a minimal network of 3 neurons, and how these dynamics can affect learning if there are global bursts. Then we discuss a 100-neurons network with global bursting suppression.

\subsection{Dynamics of LSA in a Minimal Network}
\label{sec:LSAminimal}

In \cite{LanaECAL2015} we showed that a minimal network of 2 neurons consistently follows the principle of LSA; we also showed that a single neuron is able to prune one synapse and enhance another synapse simultaneously depending on the stimulation received by the two presynaptic neurons.
In this experiment we examine the weights dynamics in a chain of 3 excitatory neurons all connected to each other, as a simplification of what may be happening in a fully connected network: one neuron is used as input, one as output, and they are separated by a ``hidden neuron".

\begin{figure}
\includegraphics[width=1.0\linewidth]{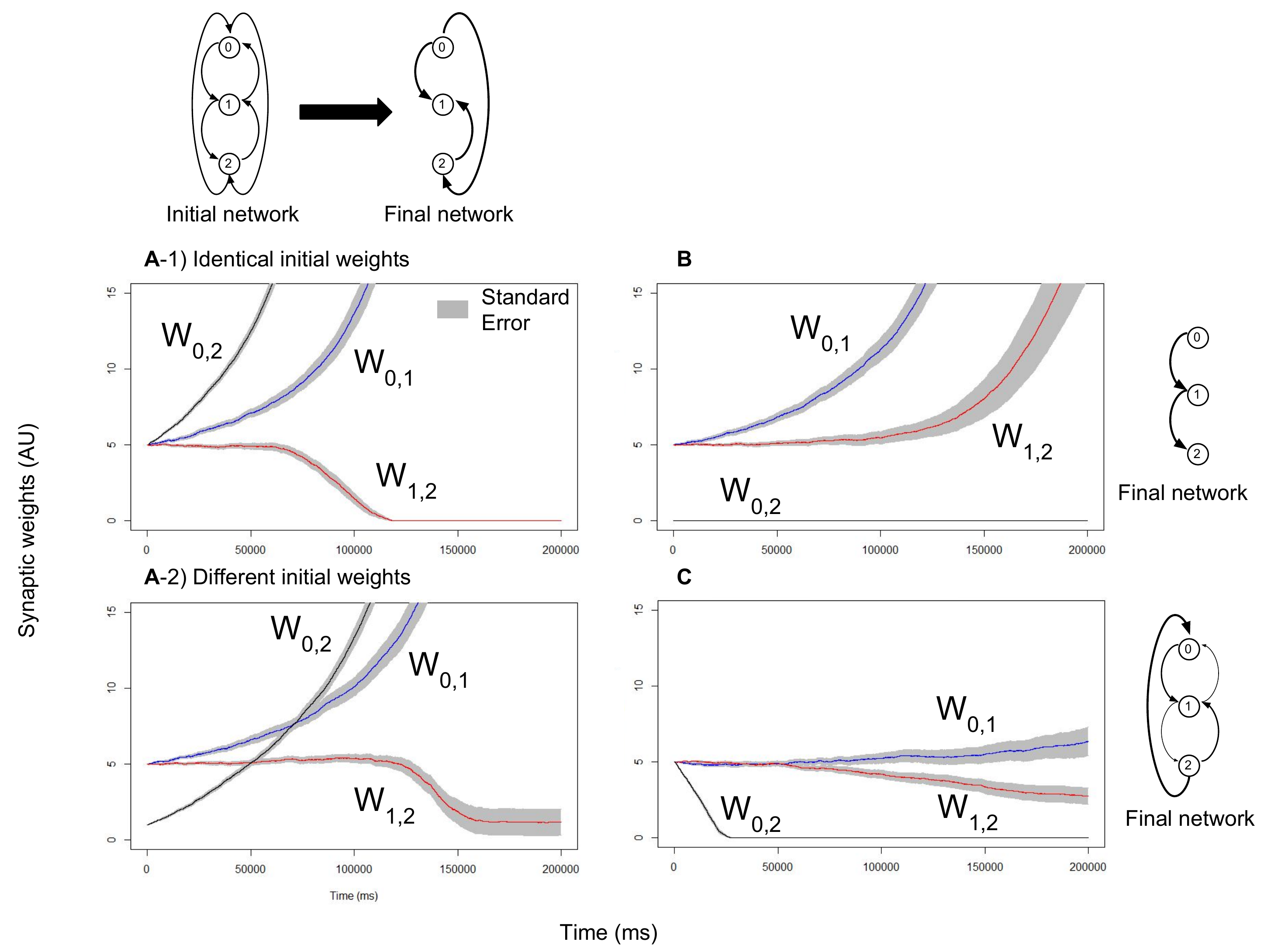}
\caption{{\bf Dynamics of weight changes induced by LSA in small networks of 3 neurons.} A) Positive LSA: spiking of 2 stops the stimulation in 0. The direct weight $w_{0,2}$ grows faster than other weights, even when starting at a lower value. B)~Artificially fixing the direct weight $w_{0,2}$ to 0, a longer pathway of 2 connections 0 $\rightarrow$ 1 $\rightarrow$ 2 is established. C) Negative LSA: spiking of 2 starts external stimulation to 0.  As a result, $w_{0,2}$ is pruned.}
\label{fig:e1v1}
\end{figure}

Neurons are labeled 0 (input neuron), 1 (hidden neuron) and 2 (output neuron). Fig~\ref{fig:e1v1} shows the results of experiments with different learning conditions and different initial states. The results can be summarised as follows:
1. In positive LSA, direct connections between input and output are privileged over indirect connections. All connections are updated with the same time step (1~ms), therefore the fastest path (direct connection) will always cause neuron 2 to fire before the longer path (made of several connections) can be completely activated. On the other hand, when no direct connection exists, weights on longer paths are correctly increased.
2. Negative LSA only prunes weights of direct connections between the input and output, as this is sufficient to stop all stimulation to the output neuron.
3. For neurons that are strongly stimulated (here, neuron 0) the default behaviour of individual weights is to increase, except if submitted to the influence of negative LSA. Neurons that fire constantly bias other neurons to fire after them, mechanically increasing their output weights.

This raises concerns about the stability of larger, fully connected networks; all weights could simply increase to the maximum value. But introducing inhibitory neurons in the network can improve network stability \cite{brunel2000dynamics}. In our experiments with 100-neuron networks, 20 are inhibitory neurons with fixed input weights and output weights.
In addition, we make the hypothesis that global bursts in the network can impair LSA, as all neurons fire together make it impossible to tease apart individual neuron's contributions to the postsynaptic neuron's excitation. Global bursts are also considered to be a pathologic behaviour for in vitro networks, and do not occur with healthy in vivo networks \cite{Bursts}.

\subsection{LSA has More Explanatory Power than the SRP}
\label{sec:LSAvsSRP}
We have seen that our simple model can reproduce the results of Shahaf's experiments, and that LSA can potentially explain this behaviour. But LSA also explains a behaviour that is not discussed in the SRP: synapse pruning. LSA predicts that networks evolve as much as possible towards dynamical states that cause the less external stimulation.

\begin{figure}
\includegraphics[width=1.0\linewidth]{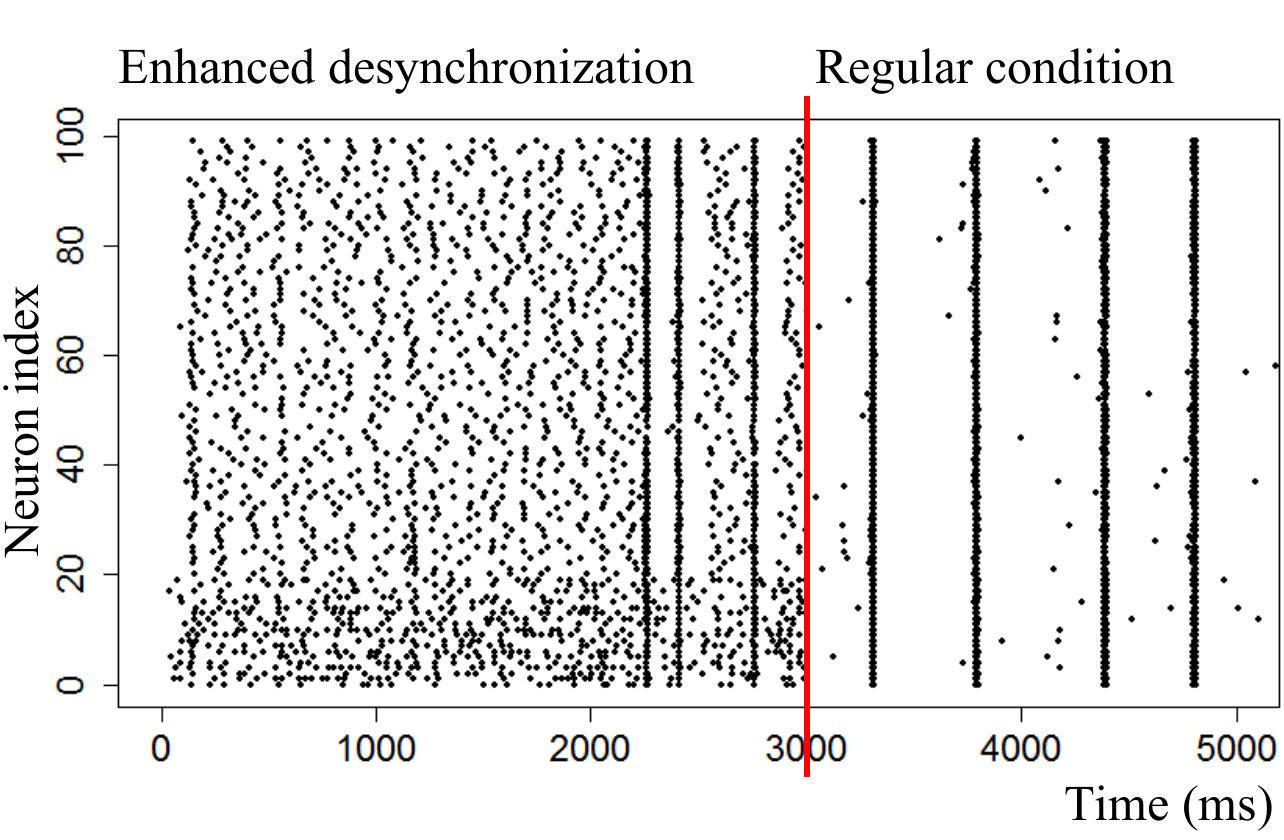}
\caption{{\bf Raster plots of a regular network (activity concentrated in bursts) and a network of which parameters have been tuned to reduce bursting and enhance desynchronized spiking.} The desynchronizing effect of sparsely connecting the network and increasing the noise are clearly visible.}
\label{fig:raster}
\end{figure}

In this experiment, we change the network's parameters to have a sparsely connected network with strong noise (see Methods). These networks are less prone to global bursts and exhibit strong desynchronized activity, as shown in Fig.~\ref{fig:raster}. 

Here we monitor two output zones and fix two stimulation conditions: (1) Input Zone A is stimulated. When a neuron fires in Output Zone A, the external stimulation to Input Zone A is stopped. (2) Input Zone B is not stimulated. When a neuron fires in Output Zone B, the whole network (excluding inhibitory neurons and Output Zone B itself) is stimulated for 10~ms. The goal is to obtain true selective learning, by increasing the firing rate of Output Zone A compared to Output Zone B.

Due to the maximum weight value, only a few (comparatively to the network size) spiking presynaptic neurons are necessary to make a postsynaptic neuron fire. In consequence, condition (2) must be able to prune as many input synapses to Output Zone B as possible. The importance of suppressing global bursts becomes obvious: global bursts cause Output Zone B to fire at the same time as the whole network, making it impossible to update only relevant weights without also updating unrelated weights.

\begin{figure}
\includegraphics[width=1.0\linewidth]{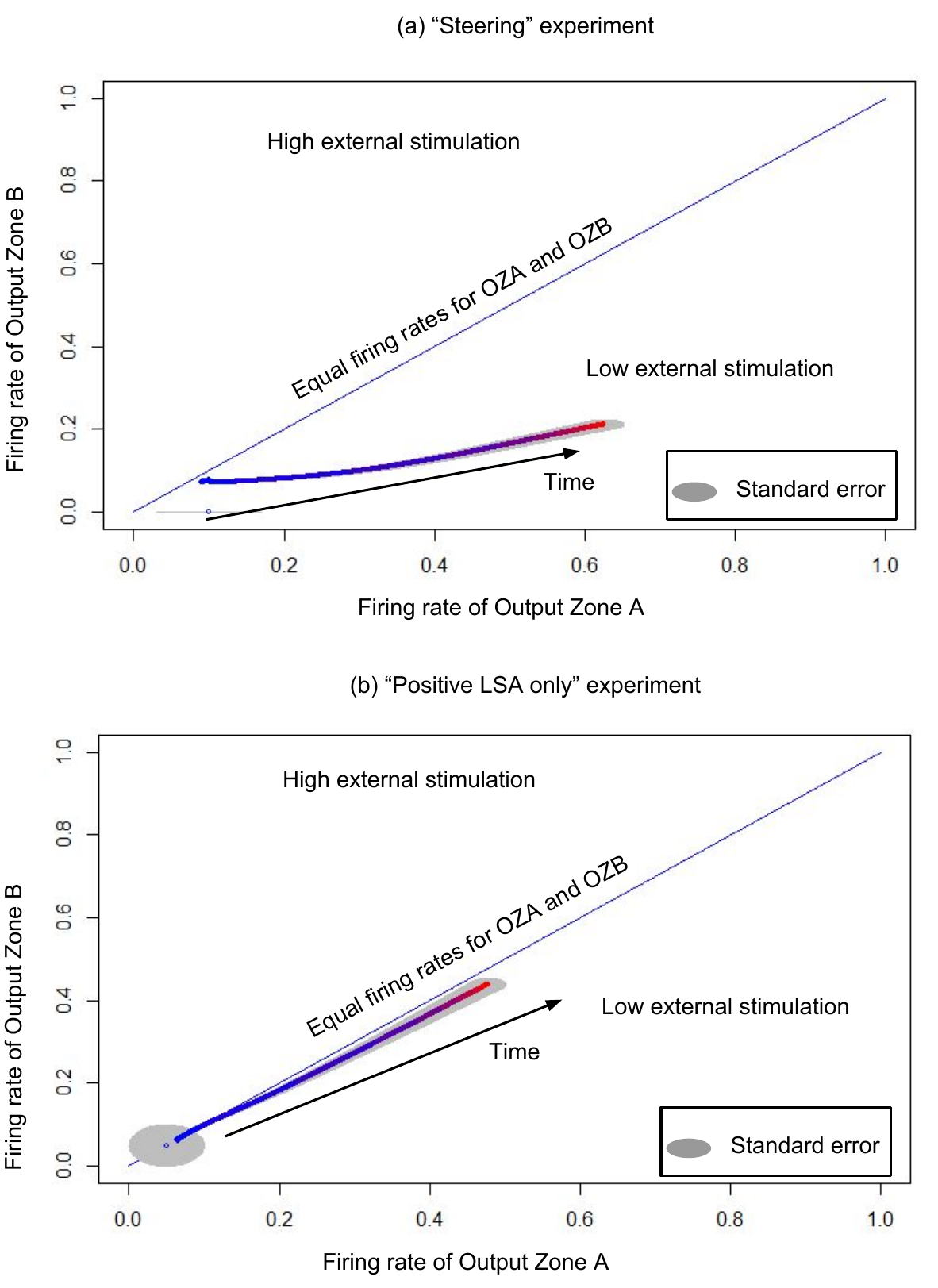}
\caption{{\bf Trajectory of the network in the two-dimensional space of the firing rates of Output Zone A and Output Zone B.} Using LSA, we can steer the network in this space; by contrast, the ``selective learning'' experiment maintains the network balanced relatively to the two firing rates.  Statistical results for N= 20 networks.}
\label{fig:steering}
\end{figure}

As a result of LSA, we expect that the network will move from a state where both output zones fire at the same rate, to a state where Output Zone A fires at high rates and Output Zone B fires at lower rates. This prediction is realized, as we can see in Fig.~\ref{fig:steering}-a: the trajectory of firing rates goes to the space of low external stimulation.  In Fig.~\ref{fig:steering}-b, we show for comparison the trajectory for networks with only condition (1) applied: on average the firing rates of both output zones are equivalent, with individual networks trajectories ending up indiscriminately at the top left or bottom right of the space.

These results could potentially be reproduced in a network in vitro: the Izhikevich model of spiking network that we use has been found to exhibit the same dynamics as real neurons, and our experiments with STDP can reproduce some results of biological experiments; therefore there is a probability that this results predicted by LSA still holds in biological networks with suppressed bursts, especially since we have shown that LSA gives promising results on biological networks embodied in simple robots (see~\cite{MasumoriECAL2015} and following section).

\subsection{Embodied Application: Wall Avoidance with a Robot} 
\label{sec:LSArobot}

\begin{figure}
\center{\includegraphics[width=0.7\linewidth]{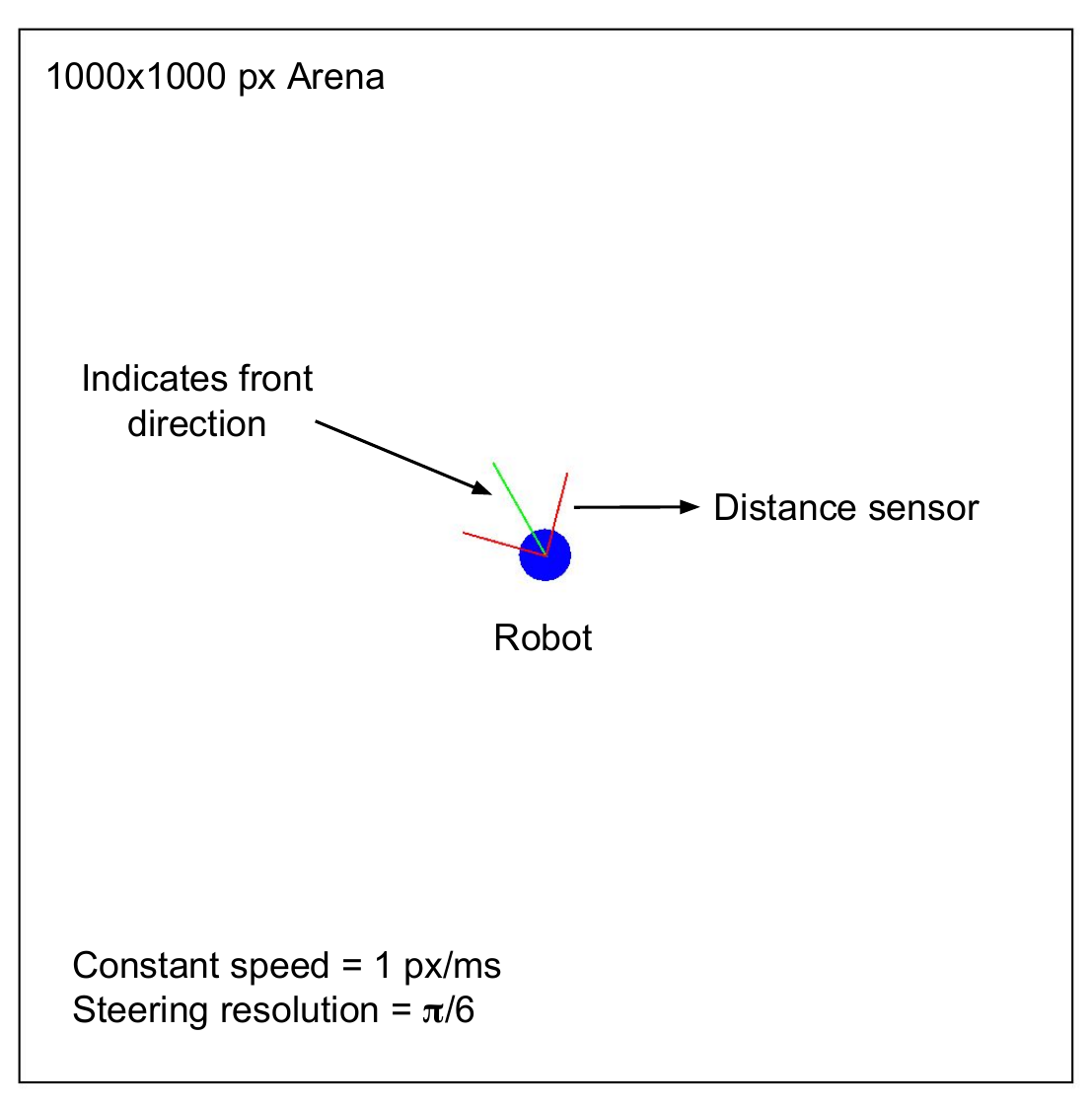}}
\caption{{\bf Robot simulation.} The robot has distance sensors and must learn to stay away from the arena's walls.}
\label{fig:robot}
\end{figure}

We show that LSA can be used in a practical embodied application: wall avoidance learning. We simulate a simple robot moving inside a closed arena. The robot has two distance sensors on the front, allowing it to detect walls (Fig.~\ref{fig:robot}). A 100-neuron network takes the two sensors' values as respective input for two input zones (right and left). Activity in two output zones of the network allows the robot to turn right or left. In these conditions, steering when encountering a wall can stop the stimulation received by the input zones from the distance sensors, if the new direction of the robot points away from the walls. The behaviour enhanced by LSA should therefore be wall avoidance. We call this experiment a ``closed loop" experiment, because steering away from the walls automatically stops the external stimulation at the right timing. 

\begin{figure}
\center{\includegraphics[width=0.8\linewidth]{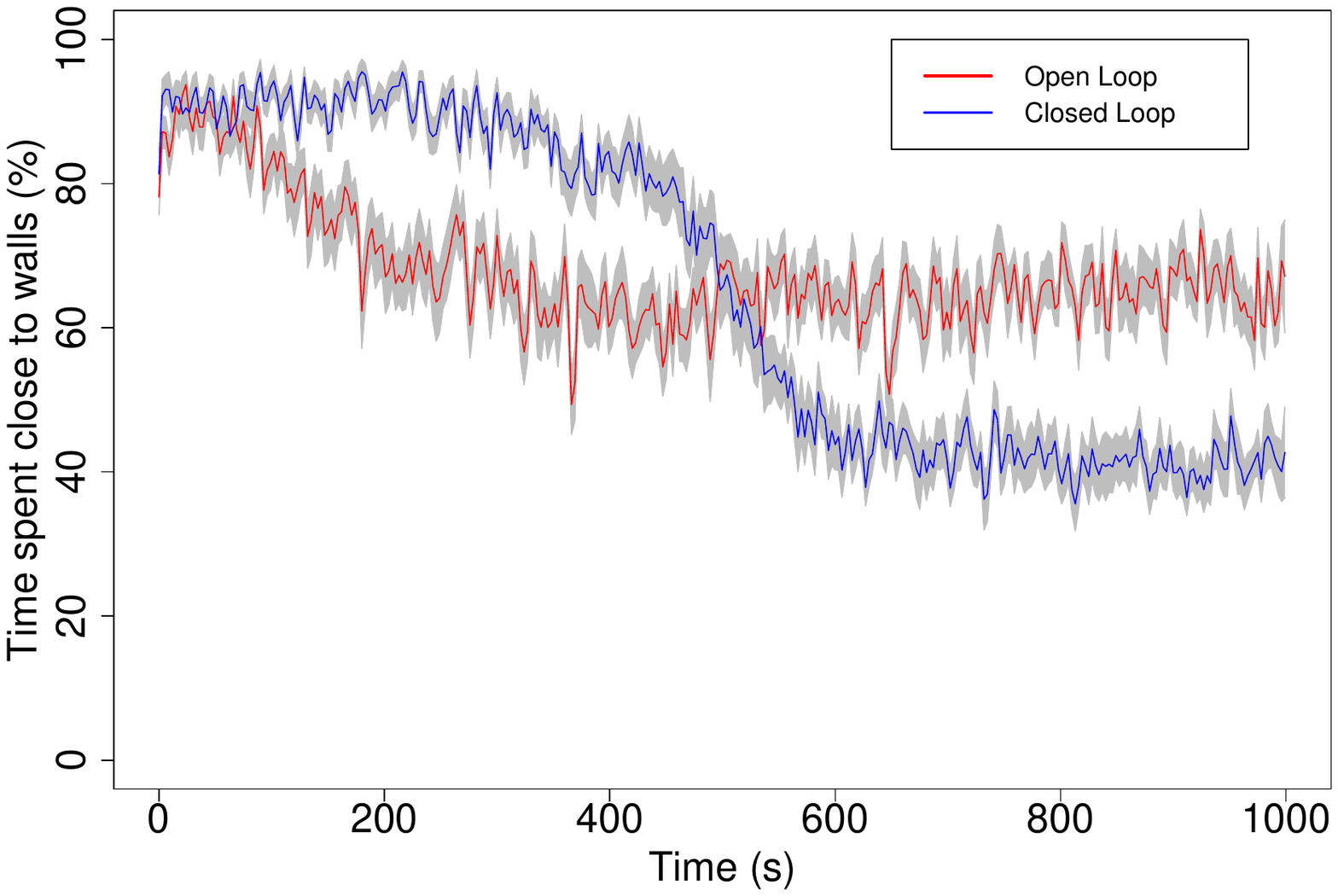}}
\caption{{\bf Learning curves of the wall avoidance task.} The robot is considered to be ``close" to a wall if it is less than 80 pixels from the wall, which corresponds to the range of its distance sensors. The results show that random steering due to high random activity in the network leads to spending 64\% of the time close to walls, while learning due to LSA leads to only 43\% time spent close to walls. Statistical results for N= 20 networks, standard error is indicated.}
\label{fig:wall_learning}
\end{figure}

In this network, burst suppression is mediated via STP (see Methods). We compare the results of this experiment (closed loop experiment, sensor $sensitivity=8$~mV) with a control experiment where a constant stimulation (8~mV) is applied to the network's input zones, independently of the distance or orientation of the robot relative to the walls (open loop experiment). Both conditions are tested 20 times and averaged. Fig.~\ref{fig:wall_learning} shows two important effects: (1) Constant stimulation leads to higher activity in the network, provoking random steering of the robot which leads to some level of wall avoidance, but (2) Closed loop feedback is necessary to obtain actual wall avoidance learning. Indeed, by the end of the closed loop experiment the robot spends only 43\% of its time at less than 80 pixels from any wall (the range of the distance sensors), against 64\% in the open loop experiment. The importance of feedback over simply having high stimulation is further demonstrated by the fact that the open loop robot is receiving overall a greater amount of stimulation than the closed loop robot. At 400~s, when the learning curve of the closed loop robot starts sloping down, the robot has received on average 1.54~mV of stimulation per millisecond. The open loop robot, which by that time has reached its best performance, has received 16~mV/ms. 
In addition, the more the robot learns to avoid walls, the less stimulation it receives on average. Despite this we reach a final state where the robot spends most of its time far from the walls.
In a different experiment, we  give to open loop robots the same amount of average stimulation that is received by closed loop robots; by the end of the experiment (1000~s) the open loop robots still spend more than 80\% of the time close to arena walls.

\begin{figure}
\center{\includegraphics[width=1.0\linewidth]{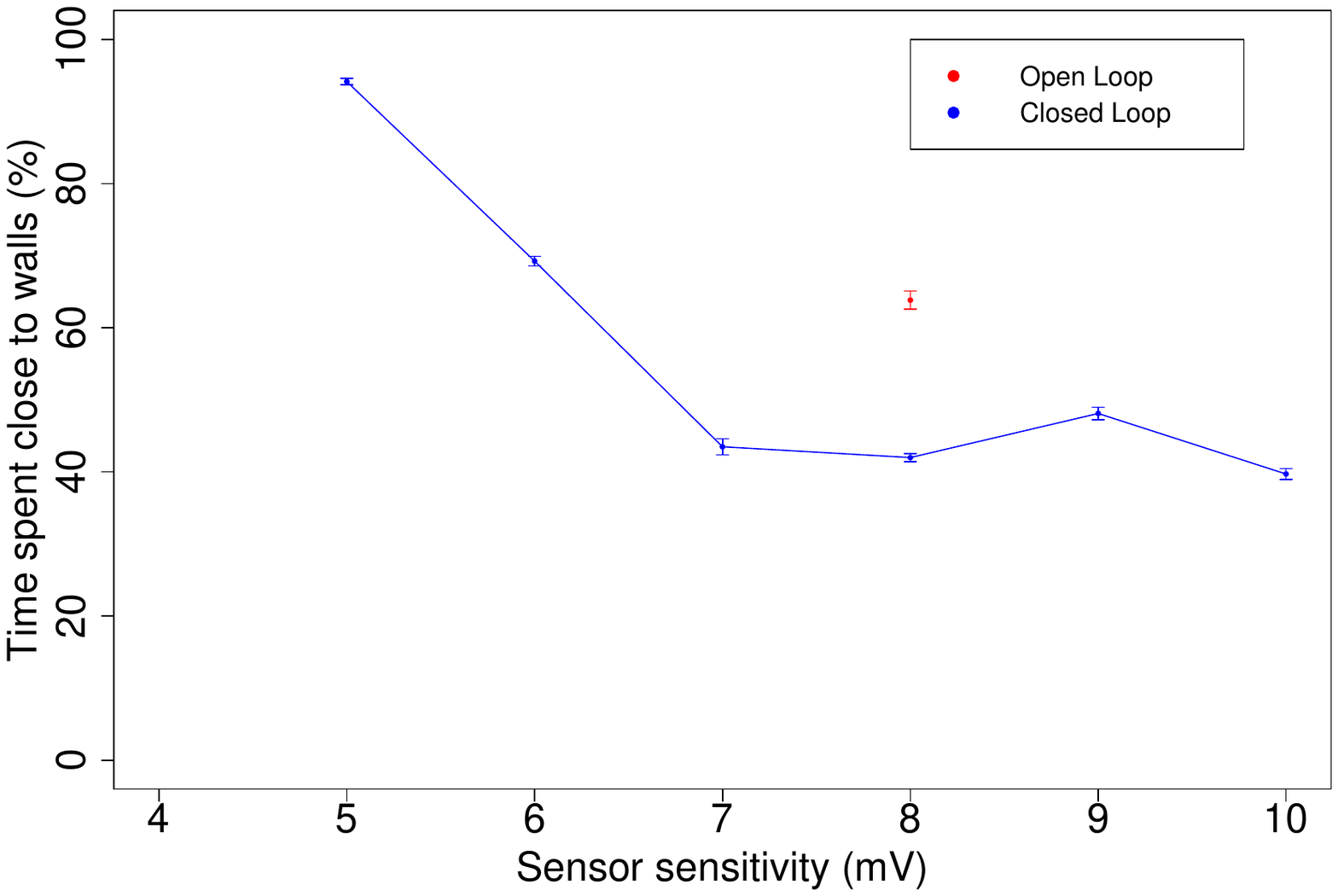}}
\caption{{\bf Learnability of wall avoidance based on sensor sensitivity.} Under 5~mV, the task cannot be learned; learnability improves until 7~mV, leading to the maximum performance of the robot. The open loop and closed loop results for the same sensitivity of 8~mV are reported from Fig.~\ref{fig:wall_learning}. Statistical results for N= 20 networks with standard error.}
\label{fig:learnability}
\end{figure}

We further study the effect of stimulation strength and feedback on learning performance by varying the sensitivity of the distance sensors. The average state in the last 300 seconds of each task (total duration: 1000~s) is reported on Fig.~\ref{fig:learnability}. The open loop result of the previous experiment is included for reference. Fig.~\ref{fig:learnability} indicates that the learnability of the task is improved by having more sensitive sensors, up to a limit of about 40\%. Having sensors with a sensitivity of more than 7~mV does not improve the performance of the robot. This result is in direct contradiction with the SRP's leading hypothesis, which postulates that the intensity of stimulation is the driving force behind network modification. If that was the case, more sensitive sensors should always lead to better learnability. By contrast, LSA emphasises timing, not strength of the simulation.
We hypothesise that the 40\% limit is due at least in part to unreliable feedback, as steering away from a wall can put the robot directly in contact with another wall if it is stuck in a corner: the same action can lead to start or removal of stimulation depending on the context.

\subsection{Parameter Exploration} 
\label{sec:Param}

Finally, we perform a parameter search to explore the working conditions of the ``simple learning" task. In the previous experiments, the network was fully connected and the initial connection weights followed a uniform distribution between 0 and 5 for the excitatory neurons and -5 and 0 for the inhibitory neurons.
In this section, we vary the number of connections in the network and the variance $v$ of the weights. For each neuron an output connection is chosen at random and the weight is initialised at ${w =  5 + \omega}$~(${w =  -5 + \omega}$),  with $\omega$ following a uniform distribution between $-v$ and $v$. This process is repeated $M$ times for each neuron, $0<M<150$. The same connection can be chosen twice at random, so the actual number of connections can be inferior to $M$.

\begin{figure}
\includegraphics[width=1.0\linewidth]{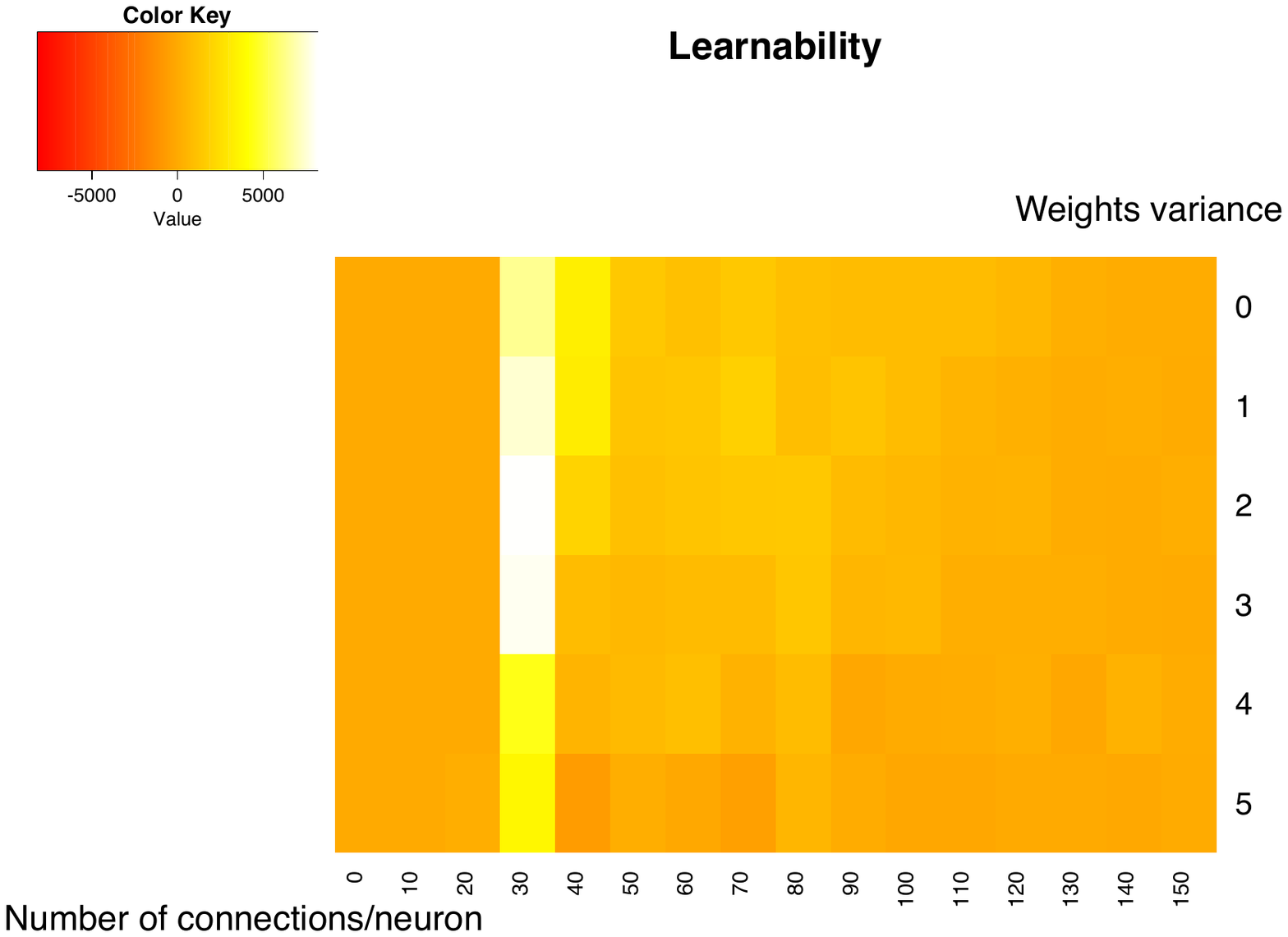}
\caption{{\bf Performance of learning depending of network connectivity and initial weights variance.} The ideal region of the parameter space to obtain good learning results is between 20 and 30 connections per neuron. By comparison, the variance has less influence. Statistical results for N= 20 networks for each parameter set.}
\label{fig:param}
\end{figure}

For each set ($\omega,M$) we perform N=20 experiments (``simple learning" experiment) of length T=500 seconds. The average difference between the firing rate of the Output Zone during the first 100 seconds and the last 100 seconds is reported on the heat map Fig.~\ref{fig:param}. This figure shows that the ideal region of the parameter space to obtain good learning results is between 20 and 30 connections per neuron. Below these values, we hypothesise that the lowest number of neurons connecting the input neurons to the output neurons becomes too big, weakening the correlation between input and output. Above these values, the increased connectivity of the network might cause too many bursts, affecting the learning results.
By comparison, the variance in the initial weights has low influence on the final learning results. 

\section*{Discussion}

In this paper, we introduce a new principle explaining the dynamics of spiking neural networks under the influence of external stimulation. The model presented in this paper is very simplified compared to biological neurons, as it uses only two types of neurons, one type of STDP, no homeostatic mechanism, etc. Nevertheless, the model is able to reproduce key features of experiments conducted biological neurons and and explain results obtained in vitro with neurons submitted to external stimulation. LSA also offers an explanation to a biological mechanism that is ignored by the theory of SRP, namely the pruning of synapses. LSA has direct practical applications: by engineering causal relationships between neural dynamics and external stimulations, we can induce learning and change the dynamics of the neurons from the outside. 

LSA relies on the mechanism of STDP, and we demonstrated that the conditions to obtain LSA are: (1) External stimulation above a minimal threshold; (2) Causal coupling between neural network's behaviour and environmental stimulation; (3) Burst suppression. We obtain burst suppression by increasing the input noise in the model or by using STP. We assume that in healthy biological neurons, the neuronal noise may be introduced by spontaneous neuronal activity.
As we have shown, LSA does not support the theory of the Stimulus Regulation Principle. It could be closer to the Principle of Free Energy Minimisation introduced by Friston \cite{friston2010free}. The Free Energy Principle states that networks strive to avoid surprising inputs by learning to predict external stimulation. An expected behaviour of networks obeying the Free Energy Principle, or obeying LSA is that they can fall into the dark room paradox, avoiding incoming input by cutting all sources of external simulation. A key difference between LSA and the Free Energy Principle is that our network does not predict incoming input. Most importantly, LSA automatically let stimuli from environment terminate at the right timing, so that a network can self-organize using environmental information.

\section*{Acknowledgments}
This work was supported by Grant-in-Aid for Scientific Research (Studies on Homeo-Dynamics with Cultivated Neural Circuits and Embodied Artificial Neural Net- works; 24300080).


%
%
%

\bibliographystyle{plos2015}
\bibliography{plos_lana_v3_arxiv}

\end{document}